\documentclass[journal]{IEEEtran}

\usepackage{multirow}
\usepackage[skip=10pt]{caption}
\usepackage{booktabs}

\usepackage{times}
\usepackage{epsfig}
\usepackage{graphicx}

\usepackage{amsmath}
\usepackage{amssymb}
\usepackage{color}
\usepackage[normalem]{ulem}
\usepackage{cite}

\usepackage{url}

\usepackage{hyperref}

\begin{document}

\title{Rate-Accuracy Trade-Off In Video Classification With Deep Convolutional
Neural Networks}

\author{Mohammad Jubran, Alhabib Abbas, Aaron Chadha and Yiannis Andreopoulos,~\IEEEmembership{Senior Member,~IEEE}
\thanks{ MJ is with the Dept. of Electrical  \& Computer Engineering, Birzeit University, West Bank, Palestine. AA and AC are with the Electronic and Electrical Engineering Department, University College London, Roberts Building, Torrington Place, London, WC1E 7JE, UK (e-mail:
\{alhabib.abbas.13, aaron.chadha.14\}@ucl.ac.uk). YA is with the Electronic and Electrical Engineering Department, University College London, Roberts Building, Torrington Place, London, WC1E 7JE, UK, and also with iSize Ltd., 41 Corsham Street, London, N1 6DR, UK, \href{www.isize.co\#isize}{www.isize.co} (e-mail:\ yiannis@isize.co). We acknowledge support from: the Leverhulme Trust (RAEng/Leverhulme Senior Research Fellowship of Y. Andreopoulos) and the Royal Commission for the Exhibition of 1851 (Fellowship of A. Chadha). MJ performed the work while visiting University College London under a ``Distinguished Scholar Award'' from the Arab Fund Fellowships programme. This work has been presented in part at the 2018 IEEE Int. Conf. on Image Process. (ICIP), Athens, Greece. }}

\markboth{IEEE Transactions on Circuits and Systems for Video Technology, to appear}
{Shell \MakeLowercase{\textit{et al.}}: Bare Demo of IEEEtran.cls for IEEE Journals}

\maketitle

\begin{abstract}
Advanced video classification systems decode video frames to derive the necessary
 texture and motion representations for ingestion and analysis by spatio-temporal
 deep convolutional neural networks (CNNs). However, when considering visual
Internet-of-Things applications, surveillance systems and semantic crawlers
of large video repositories, the video capture and the CNN-based
semantic analysis parts do not tend to be co-located. This necessitates the
transport of compressed video over networks and incurs significant overhead
in bandwidth and energy consumption, thereby significantly undermining the
deployment potential of such systems. In this paper, we investigate the trade-off
between the encoding bitrate and the achievable accuracy of CNN-based video
classification models that directly ingest  AVC/H.264 and HEVC encoded videos. Instead of retaining entire
compressed video bitstreams and applying complex optical flow calculations prior to CNN processing, we only retain motion vector and select texture
information at significantly-reduced bitrates and apply no additional processing prior to CNN ingestion. Based on three  CNN architectures
and two action recognition datasets, we achieve 11\%--94\%\  saving in bitrate
with marginal effect on classification accuracy. A model-based selection
between  multiple CNNs increases these savings further, to the point where, if up to 7\% loss of   accuracy can be tolerated, video classification\ can take place with as little as 3 kbps  for the transport of the required compressed video information to the system implementing the CNN  models.
\end{abstract}
\section{Introduction}
\label{sec:intro}

Action or event recognition and video classification for visual Internet
of Things (IoT) systems \cite{posada2015visual,redondi2012rate,redondi2013compress}, video surveillance \cite{yu2015fast},
and fast analysis of large-scale video libraries \cite{abu2016youtube} have
been advancing rapidly due to the advent of deep convolutional neural networks
(CNNs). Given that such CNNs are very computationally and memory intensive,
they are not commonly deployed at the video sensing nodes of the
system (a.k.a., ``edge'' nodes). Instead, video is either transported to
certain high-performance aggregator nodes in the network  \cite{posada2015visual,redondi2012rate,redondi2013compress}
that carry out the CNN-based processing, or compact features are precomputed
in order to allow for less complex on-board processing at the edge \cite{abu2016youtube},
usually at the expense of some accuracy loss for the classification or recognition
task.

Motion vector based optical flow approximations 
have been proposed for action recognition by Kantorov and Laptev \cite{mpegflow},
albeit without the use of CNNs. In more recent work, proposals have been
put forward for fast video classification based on CNNs that ingest compressed-domain
motion vectors and selective RGB\ texture information  \cite{chadha2017video,zhang2016real}.
Despite their significant speed and accuracy improvements, none of these
approaches considered the trade-off between rate and classification accuracy
obtained from a CNN. Conversely, while rate-accuracy trade-offs have been
analysed for conventional image and video spatial feature extraction systems \cite{redondi2012rate,redondi2013compress},
these studies do not cover deep CNNs and semantic video classification, where
the different nature of the spatio-temporal classifiers can lead to different
rate-accuracy trade-offs.

In this paper, we show that crawling and classification of remote video data
can be achieved with significantly-reduced bitrates by exploring
 rate-accuracy trade-offs in CNN-based classification. Our contributions are summarized as follows:

\begin{enumerate}

\item We study the effect of varying   encoding parameters on state-of-the-art CNN-based video classifiers. Unlike conventional  rate-distortion  curves,
we show that, without any optimization,  rate-accuracy is not monotonic for CNN-based classification.

\item In order to
 optimize the trade off between bitrate and classification accuracy, we propose a  mechanism to select amongst 2D/3D temporal CNN and spatial CNN classifiers that have varied input  volume requirements. We achieve this\ with minimal  modifications to the encoded bitstream, which are straightforward to implement in practice.
\item We study and compare the efficacy of our method on action recognition based on AVC/H.264 and HEVC compressed video, which represent two of the  most commonly-used video coding standards.   
\end{enumerate}
  
These contributions extend on our recent conference paper on this subject \cite{abbasICIP}, which did not cover the last two points from above. The remainder of the paper is organized as follows. In Section \ref{sec:related_work}, we give an overview of recent work on compressed video classification. Section  \ref{sec:highly_compressed_bitstreams} details how we reduce video bitstreams through selective cropping. In  Section \ref{sec:network_design}, we describe and formulate the optimized classifier selection process. Section \ref{sec:Evaluation} evaluates the performance of the proposed classifiers using different coding settings and illustrates the rate gains made possible through our classifier selection method. Finally, Section \ref{sec:Conclusion} concludes the paper.

\section{Related Work \label{sec:related_work}}

The use of codec motion vectors as an approximation of optical flow 
has  been proposed for action recognition by Kantorov and Laptev \cite{mpegflow}.
Their approach preceded the surge in convolutional neural networks for image
classification and used Fisher vectors, which achieve lower accuracies
in   standard action recognition datasets. More recently, Zhang \textit{et
al.}
\cite{zhang2016real}  utilized codec motion vectors as input to a 2D CNN
for
action recognition with a  framework that requires optical-flow
based training and transfer learning \cite{pan2010survey}. Their
requirement of highly-upsampled frames  during inference increases the implementation complexity,  as large
activation maps need to be calculated at the first layers of their  CNN. Recent work \cite{chadha2017video,wu2017compressed} showed that compressed-domain action recognition \ can achieve accuracy that competes with optical-flow based methods, while offering higher ingestion and CNN\ processing speed than all previous alternatives. 
 Given that  the spatial stream  learns on scene
information that tends to be persistent  across frames,  compressed-domain methods gain
by  sparse
frame decoding combined with motion-adaptive  super-positioning of  decoded
macroblock information to generate   intermediate frames at a finer
temporal
scale.

However, thus far, there has been no work on exploring rate-accuracy tradeoffs for CNN-based video classification. This is now increasingly important due to the advent of visual IoT and cloud-based platforms, where the visual sensing and processing are not co-located  \cite{posada2015visual,redondi2012rate,redondi2013compress}. Alas, such tradeoffs are non trivial, because they depend on the spatio-temporal information needed by the CNN performing the recognition task \cite{girdhar2017actionvlad,girdhar2017attentional}. For instance, one of the  issues with most of the work described above is the short
temporal extent of  inputs \cite{chadha2017video,xu2017two,zhao2017pooling}; each input
video segment comprises a small group of frames that only represent (approximately) one second of the recorded action or event to be classified.
 Hence, this cannot account for cases where temporal dependencies extend over
longer durations  \cite{chadha2017video}.  Feichtenhofer \textit{et al.} \cite{feichtenhofer2016convolutional}
attempted to resolve this issue by using multiple copies of their two stream
network where the copies are spread over a coarse temporal scale, thus encompassing
both coarse and fine motion information with an optical flow input.  The
 architecture is spatially and then temporally fused   using 3D convolution
and pooling. Despite achieving state-of-the-art results on UCF-101 and HMDB-51
datasets, this approach  requires heavy processing for both training and
testing.  Alternatively, other work \cite{girdhar2017actionvlad,varol2016long} argues
that increasing the temporal extent is simply a case of taking the optical
flow component  over a larger temporal extent.  In order to minimize the
complexity of the network, most such approaches downsize the frames,  thus reducing the spatial
dimensions.  On the other hand, the work of Sevilla \textit{et al.}  \cite{sevilla2017integration}  shows that   high-resolution
 optical flow can be beneficial  since deep learning methods can learn features from small details.
This observation suggests that high-resolution   optical flow  can be leveraged to lower the temporal extent of inputs. Understanding the trade-offs in  compressed-domain spatio-temporal information and exploring the rate-accuracy characteristics of CNN-based video classification  is the objective of this paper.

\section{Cropped Video Bitstreams} \label{sec:highly_compressed_bitstreams}

We base our reduced-bitstream encoder on the JM reference software  of  AVC/H.264
\cite{alexis2009jm}  and the HM\ reference software of HEVC   \cite{H264Std}.
Our modifications to the reference encoders are designed such that the bitrate of the compressed
bitstream is kept at a minimum while preserving the information needed to
classify videos. Namely, the compressed bitstream  should exclusively hold:
\textit{(i)} key texture components corresponding to rapidly-changing input content;  \textit{(ii)} inter-frame predicted macroblocks and their  motion compensation parameters; \textit{(iii)} control signals
and headers needed to comply with its corresponding standard. 
\subsection{Summary of Spatio-Temporal Representations in Video Coding Standards }
Before applying inter-frame prediction,  AVC/H.264 pictures are split into
$16 \times 16$ pixel macroblocks (MB)
to represent luminance and chrominance samples,  with the chrominance samples
further split into  $8\times8$ chroma
blocks for the widely used $4$:$2$:$0$ chroma sampling. Macroblocks are the
core of the coding
layer and form the basis for  adaptive inter and intra predictions. Each
of the inter-predicted macroblocks is then encoded using  blocks from the
 set
$\{16\times16, 16\times8, 8\times16,8\times8\}$ \cite{H264Std,wiegand2003overview}.
The  HEVC standard takes on a more adaptive approach and introduces a  Coding
 Tree Unit (CTU) which consists
of luma and chroma Coding Tree Blocks (CTB). 
The size of each luma CTB is drawn from the set $\{16\times16, 32\times32,
64\times64\}$ where larger size blocks result
in better compression efficiency.   Iterative
partitioning is then applied to divide  CTBs into smaller Coding Blocks (CB)
resulting in a tree-like structure
 \cite{Samet1984Quadtree}. The minimum allowed CB size is also specified,
this serves as a hyper-parameter to  control  the granularity of the  tree
structure produced,  this parameter is  commonly referred to as \textit{depth}
 \cite{Sullivan2012HEVC}.

In both standards, blocks are predicted via translational motion vectors
(MVs) that represent
the displacement from matching blocks in previous or subsequent reference
frames. Increasing the number of small-size blocks increases the granularity
of the MV grid at the expense of lower coding efficiency. These MVs represent
the temporal activity and have been shown to be highly correlated with optical
flow estimates   \cite{chadha2017video}. If the area covered by the MB is
static, the MB is ``skipped'' and is not encoded. The resulting prediction
residual from temporal prediction of  non-skipped MBs is encoded using transform
coding. The transform coefficients are then quantized based on the quantization
parameter (QP). The value of the QP per frame can be chosen from  52 values
in $[0,51]$, with lower values indicating high-fidelity\   encoding.

\subsection{Selective Retention of Motion and Texture Information}

\begin{figure*}[ht!]    \includegraphics[scale=0.57]{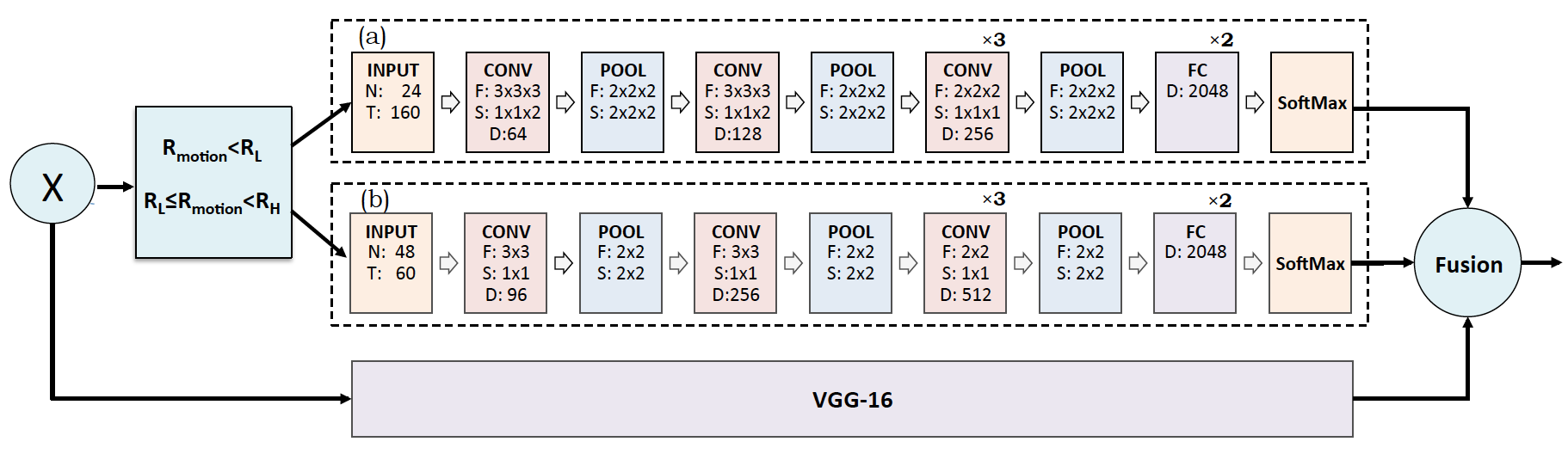} \caption{\label{fig:mcnn} The proposed Multi-CNN classifier selection: (a) 3D temporal CNN architecture; (b) 2D temporal CNN architecture. The bottom part represents the spatial CNN (VGG-16). Parameters: N is the spatial dimensions of the  input volume;  T is the temporal extent expressed as  the  number of frames used; F is the filter size, formatted as width $\times$ height $\times$ time; S is the convolutional window stride; D is the number of filters (or number of hidden units) for the convolutional and fully-connected layers; $R_\text{L}$ and $R_\text{H}$ are controlling the multi-CNN selection based on the motion vector rate $R_\text{motion}$. } \end{figure*}


In our work, only select subsets of the quantized transform coefficients
will be entropy encoded and then included in the cropped bitstream. This
set of coefficients, along with spatial texture, is   transmitted to the classifier  (described in Section \ref{sec:network_design}) to infer semantic features and classify
the content of the bitstream. By doing so, the  bitrate of these ``cropped'' subsets of coefficients,  $R_{\text{cropped}}$, is significantly reduced in comparison to the original bitrate,  $R_{\text{orig}}$, needed to encode the full video.  In the remainder, we present our modifications, assuming that the first frame of every video sequence
is encoded as an Instantaneous Decoding Refresh (IDR) and all subsequent
frames in the video are encoded as P-frames. 

In order to reduce the bitrate of the compressed bitstream, we  employ selective
retention
of texture information by retaining the texture information of active regions.
 To implement selective
writing in the AVC/H.264 JM reference software \cite{alexis2009jm}, we modified functions \texttt{writeCoeff4x4\_CAVLC\_normal()}
and\   \texttt{write\_chroma\_intra\_pred\_mode()}. In addition, to allow for a skip symbol for all non-active areas, we modified the functions \texttt{read\_coeff\_4x4\_CAVLC()} and \texttt{read\_coeff\_4x4\_CAVLC\_444().}  Similarly, to implement selective writing in the HEVC HM reference software \cite{Sullivan2012HEVC}, we modified the functions \texttt{TEncSbac::codeCoeffNxN()}
 and  \texttt{TDecSbac::parseCoeffNxN().}  To simplify
our tests, we retain the texture of  IDR\ frames and  skip all texture
of P-frames with a single skip symbol. The introduction of these skip symbols is the only non-normative part of our entire process. All other syntax
elements (including modes
and motion information) are left as specified in their respective standard.
With these minimal changes,  standard decoders can decode our reduced bitstreams to pass to compressed video classifiers. 

Finally, in order to  derive a temporal activity
map from P-frame  MVs, we apply the
following steps: \textit{(i)} MVs are  extracted from the compressed bitstream
using the \texttt{read\_motion\_info\_from\_NAL\_p\_slice()} function for JM  and \texttt{TDecEntropy::decodePUWise()} for HM; \textit{(ii)} the extracted MVs are then mapped to a grid of  $8\times8$  non-overlapping blocks within each frame; \textit{(iii)} MVs are interpolated from neighboring macroblocks wherever a macroblock does not
provide  motion compensation parameters but two or more of its neighbors do.

\section{Proposed Framework For Compressed-domain Classification}
\label{sec:network_design}

\subsection{CNN Architectures}In Fig. \ref{fig:mcnn} we illustrate  the two
CNNs\  used for the temporal MV stream, which represent the state-of-the-art in compressed-domain deep learning for action classification\cite{chadha2017video}\cite{zhang2016real}. We use two architectures
 to study how 
 different models behave to cropped bitstream volumes, and to demonstrate
that
 our rate optimized CNN-based classification method is applicable with  different
 network architectures that have been shown to preform well with  codec motion vector data.
The first CNN architecture we consider is the 3D CNN proposed by Chadha \textit{et al.}   \cite{chadha2017video}.  As illustrated in Fig. \ref{fig:mcnn}(a),
all convolutional and pooling layers are spatiotemporal in extent; this
 captures the motion information between
consecutive motion vector frames.  Crucially, the spatiotemporal features
are expected to improve classification performance between similar actions.
We generate a 4D motion vector input by splitting the $dx$ and $dy$ vector
components into separate channels, thus resulting in a $W\times H\times 2\times
T$ volume.  We compensate for the low resolution of the extracted motion
vector frames by setting a long temporal extent $T$ as $T_{3D}=160$, which typically
comprises the entire video duration.

The second architecture we consider is a 2D  CNN, as illustrated in Fig.
\ref{fig:mcnn}(b). The model design is based on ClarifaiNet \cite{russakovsky2015imagenet}
and only comprises 2D spatial filters; we notably reduce the size of the
first filter from $7\times 7$ to $3\times 3$ and decrease the stride of the
first two convolutional layers to $1\times1$.  A similar architecture was
also employed in recent work on fast video classification \cite{zhang2016real}.
The input is generated by stacking the motion vector $dx$ and $dy$ components
into a single $W \times H \times 2T$ volume, where the temporal depth $T$
is set as $T_{2D}=60$.  In general, 2D CNNs are less complex to train and test with than 3D CNNs, whilst forgoing modelling any temporal dependencies. Nonetheless,
their lower complexity means we can afford to use a higher input spatial resolution, which
enables the 2D filters to learn more distinguishing spatial features of the MV data.

Finally, concerning spatial processing of RGB\ texture,
we use the well-established VGG-16 \cite{simonyan2014very}
CNN architecture to classify RGB frames and capture motion-invariant spatial
features of video content. Our spatial CNN is pre-trained on ImageNet \cite{deng2009imagenet}
and fine-tuned on the training split of UCF-101. The spatial stream ingests the decoded frames
per video and the predictions made by the spatial CNN\ are ultimately fused
with the predictions from the temporal stream to produce the final two-stream
classifier decisions.

\subsection{Training and Testing} \label{sec:training_proc}

We train both temporal stream architectures using stochastic gradient descent
with momentum set to 0.9. The initialization of He \textit{et al.}
\cite{he2015delving}   is used and  weights are initialized
from a normal
distribution. Mini-batches of size 64 are generated by randomly selecting
64 training
videos per batch.  The learning rate is initially
set to $10^{-2}$ and is decreased by a factor of $0.1$ every 30k
iterations. The training is completed after 90k iterations. We follow the  data augmentation practices utilized in recent work \cite{chadha2017video}
in order to minimize overfitting for both the 2D and 3D CNN.  These include
a multi-scale random cropping of the input and spatial resizing to a fixed
size $N$, followed by zero centering the motion vector field by subtracting
the mean motion vector from the volume. For the 3D CNN, the fixed crop size
is set to 24, whereas for the 2D CNN this is doubled to 48. In addition,
we use a dropout ratio to 0.5 for the first two fully connected layers in
both models.
During testing, for the temporal stream we generate 10 random volumes of
temporal size $F$ from which to test on. Per volume, we use the standard
10-crop testing, cropping the four corners and the center of the image to
spatial size $N \times N$ and considering both horizontally flipped and unflipped
versions. As such, we average the scores over 10 crops and 10 volumes to
produce a single score for the video. For the spatial stream, we use one IDR frame for each video and  oversample inputs to VGG-16 by flipping and extracting crops.

%
%
%

\subsection{Multi-CNN\ Classifier}\label{sec:DualTemporal}

In order to optimize the tradeoff between bitrate and classification accuracy, we leverage the differences in input requirements of the two temporal classifiers of    Fig. \ref{fig:mcnn} and devise a Multi-CNN\  (MCNN) selection process. Since the number of MV frames per crop is larger for our 3D CNN\ versus its 2D CNN
counterpart (i.e., $T_{3D}>T_{2D}$), the former requires higher bitrate per crop than the latter. On the other hand, as shown in  previous studies  \cite{sevilla2017integration}, denser
MV frames will benefit from the spatially-larger input of the proposed 2D CNN architecture. Since the density of inputs to
the temporal stream is directly  proportional to the average bitrate allocated to MVs by the codec    $R_{\text{motion}}$, we expect  the accuracy of both the 2D CNN and 3D\ CNN classifiers to be directly related to $R_{\text{motion}}$, albeit up to a limit (since noise is introduced at high rates due to the limitations of the MV block model).  Moreover, the two classifiers are expected to be comparable in accuracy  over a  range of  $R_{\text{motion}}$ values. These hypotheses have been tested and we present the related experimentally derived results  in the  Appendix. In summary, our investigation showed that: \textit{(i)} the long temporal extent 3D CNN\ classifier is superior for lower values of $R_\text{motion}$; \textit{(ii)} the short temporal extent 2D CNN\ classifier performs as well as the long temporal extent 3D CNN classifier for  mid-range values of $R_\text{motion}$ ; \textit{(iii)} both temporal CNNs offer diminishing performance for high values of $R_\text{motion}$. Therefore, we introduce the pair of rate-accuracy
optimization parameters \{$R_\text{L}$, $R_\text{H}$\}, with $R_\text{H}>R_\text{L}$,  such that: 
\begin{itemize}
\item
 the 3D CNN is used for  videos with  $R_{\text{motion}} <  R_{\text{L}}$     
\item the 2D CNN is used for  videos with $ R_{\text{L}}\leq R_{\text{motion}} <  R_{\text{H}}$
\item no temporal CNN is used when $R_\text{motion} \geq \ R_{\text{H}}$ and only the output of the spatial CNN is considered (see  Fig. \ref{fig:mcnn}).   
\end{itemize}
The remainder of this section is to establish a model-based approach for the optimal selection of  \{$R_\text{L}$, $R_\text{H}$\}. While  the value of $R_{\text{motion}}$ is  derived experimentally during
the encoding of each video, for offline rate-accuracy optimization studies
it can also be derived via rate-distortion models \cite{kwon2007rate}.    

%
%
%
%
%
%
%

\begin{figure*}[ht] \centering \includegraphics[scale=0.52]{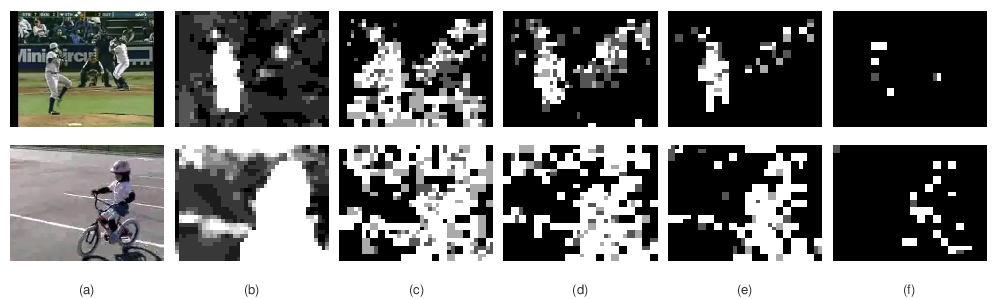}
        \caption{\label{fig:mix_sparsity} RGB frames and  corresponding
AVC/H.264 MV activity maps  for two scenes from  UCF-101; (a) RGB frames; (b)
Brox optical flow; (c) Approximated flow 
at $\text{QP}=0$; (d) Approximated flow at $\text{QP}=30$; (e) Approximated flow at $\text{QP}=40$; (f) Approximated flow at $\text{QP}=51$.  Note that sparsity increases and noise decreases with increased $\text{QP}$.} \end{figure*}

\subsection{Problem Formulation and Optimization of  MCNN}\label{sec:OptMultiCNN}

To make full use of the overlap of performance between  classifiers,  a video  is passed to  a lower-rate classifier only when it is  likely to be classified correctly. We consider the problem of finding the optimum set $\{R^*_{\text{L}}, R^*_{\text{H}}\}$
that maximizes the classification accuracy,  $A_\text{mcnn}$, of our proposed MCNN under a constraint on the available bitrate,  $R_\text{available}$:

\begin{equation}\label{eq:Maxmin}
\begin{aligned}
\{R^*_{\text{L}}, R^*_{\text{H}}\}=\underset{R_{\text{L}}, R_{\text{H}}}{\mathrm{arg}\mathrm{max}}A_{\text{mcnn}}
 \text{subject to} R_{\text{sent}}\leq R_{\text{available}}
\end{aligned}
\end{equation}where $R_\text{sent}$ is the average bitrate of all transmitted bitstreams under a selection algorithm for \{$R_\text{L}$, $R_\text{H}$\}.  We first consider the  video source probability density function $f_\text{s}(R_{\text{motion}})$, which characterizes the probability of occurrence of video examples with bitrate $R_\text{motion}$. We  have found
$f_\text{s}(R_{\text{motion}})$ to be well approximated by the
Gamma distribution,   $f^{}_\text{s}(R_\text{motion};\alpha,\beta)$, where  $\alpha$ and $\beta$ are the shape and rate parameters (see Section \ref{sec:app_fitted_model_perf_overlap} of the Appendix and     Fig. \ref{fig:rmotion_gamma}).  We can then express  $A_{\text{mcnn}}$ as: 
 
\begin{equation}\label{eq:Acc-mcnn}
\begin{aligned}
A_{\text{mcnn}}=A_{\text{3D}}\int _{0}^{R_{\text{L}}} f_\text{s}(R_{\text{motion}})dR_{\text{motion}}\\ +A_{\text{2D}}\int _{R_{\text{L}}}^{R_{\text{H}}} f_\text{s}(R_{\text{motion}})dR_{\text{motion}}\\ +A_\text{SP}\int _{R_{\text{H}}}^{\infty} f_\text{s}(R_{\text{motion}})dR_{\text{motion}}
\end{aligned}
\end{equation} 
where $A_{\text{3D}}$, $A_{\text{2D}}$ and $A_\text{SP}$ 
are the  classification accuracies of the 3D, 2D and spatial stream classifiers respectively.
In \eqref{eq:Acc-mcnn}, the  accuracy of each of the classifiers is assumed to be constant for the range of rates it corresponds to,  and its estimate is experimentally derived from $\mathcal{V}$.   This  assumption holds as long as  $\mathcal{V}$ is large enough and the accuracy of each classifier remains relatively flat for different values of $R_{\text{motion}}$ within the respective integration interval of each classifier, which is found to be the case in our experiments of Section \ref{sec:Evaluation}. 

Since   the
 number of bits needed to classify each video depends on which classifier
is used for prediction, we first find the average bitrate required by each  classifier.  We  define $R_{\text{3D}}$, $R_{\text{2D}}$, and $R_\text{SP}$ as the average bitrate of  inputs to the  3D, 2D, and spatial classifiers, respectively, and  estimate each as:

\begin{equation}\label{eq:R_3D_2D_SP}
\begin{aligned}R
_{}=\begin{cases}R_{\text{3D}}=a_{\text{3D}}R_{\text{motion}}^{}+b_{\text{3D}} & 0 \ \ <\ R_{\text{motion}}<R_\text{L}
\\
R_{\text{2D}}=a_{\text{2D}}R_{\text{motion}}^{}+b_{\text{2D}} & R_\text{L}\leqslant R_{\text{motion}}<R_\text{H}\\
R_{\text{SP}}=I_\text{SP} & R_\text{H}\leqslant R_{\text{motion}}<\infty\ \\
\end{cases}
\end{aligned}
\end{equation} 
%
%
%
where $a_{\text{3D}}$, $b_{\text{3D}}$,
$a_{\text{2D}}$, and $b_{\text{2D}}$ are coefficients to be estimated by applying
regression on the    bitrate feature $R_{\text{motion}}$ obtained on the  training set $\mathcal{V}$.  Since the inputs passed to the 3D and 2D classifiers consist only of the motion vectors and some added headers to comply with the used standard,  we expect the linear relations shown in \eqref{eq:R_3D_2D_SP} and confirm this in Section  \ref{sec:app_3d_2d} of the Appendix. For the spatial classifier, we  use $I_\text{SP}$, i.e., the bitrate of the first IDR frame, to estimate $R_\text{SP}$. Note that $R_\text{motion}$ is not used for $R_\text{SP}$, since the spatial classifier  only uses texture information. We can now express $R_{\text{sent}}$ as:  
\begin{equation}\label{eq:Rsent}
\begin{aligned}
 	{R_\text{sent}}=\int _{0}^{R_\text{L}} R_{\text{3D}}{f_\text{s}}(R_{\text{motion}})dR_{\text{motion}}\\ +\int _{R_\text{L}}^{R_\text{H}} R_{\text{2D}}{f_\text{s}}(R_{\text{motion}})dR_{\text{motion}}\\ +\int_{R_\text{H}}^{\infty} R_\text{SP}{f_\text{s}}(R_{\text{motion}})dR_{\text{motion}}
\end{aligned}
\end{equation}
Based on the expectation value property of the Gamma density function $f(X;\alpha,\beta)$ \cite{degroot2012probability}: 
\begin{equation}\label{eq:gamma_property}
\begin{aligned}
Xf(X;\alpha,\beta)=\frac{\alpha}{\beta}f({X;\alpha+1},\beta) 
\end{aligned}
\end{equation}
from \eqref{eq:Acc-mcnn} and \eqref{eq:Rsent} we can  rewrite  $A_{\text{mcnn}}$ and $R_{\text{sent}}$ as:
\begin{equation}\label{eq:Acc-mcnn-Gama}
\begin{aligned}
A_{\text{mcnn}}=(A_{\text{3D}}-A_{\text{2D}})F_\text{s}(R_\text{L};\alpha,\beta)\\ + \ (A_{\text{2D}}-A_{\text{SP}})F_\text{s}(R_\text{H};\alpha,\beta)+A_{\text{SP}}
\end{aligned}
\end{equation}

\begin{equation}\label{eq:Rbudget_Gama}
\begin{aligned}
R_{\text{sent}}=(b_{\text{3D}}-b_{\text{2D}})F_\text{s}(R_\text{L};\alpha,\beta)\\ + (b_{\text{2D}}-I_{\text{SP}})F_\text{s}(R_\text{H};\alpha,\beta)\\ +(\alpha/\beta)(a_{\text{3D}}-a_{\text{2D}})F_\text{s}(R_\text{L};\alpha+1,\beta)\\ + (\alpha/\beta)(a_{\text{2D}})F_s(R_\text{H};\alpha+1,\beta)+I_{\text{SP}}
\end{aligned}
\end{equation}

where  $F_\text{s}$ is the cumulative distribution function of  $f_\text{s}$ and we have explicitly indicated the dependence on the parameters $\alpha$ and $\beta$ since they affect the bitrate and accuracy contributions of the 2D and 3D CNN models.
The constrained optimization problem of \eqref{eq:Maxmin}
 can now be solved  for $\{R^*_{\text{L}}, R^*_{\text{H}}\}$ via  \eqref{eq:Acc-mcnn-Gama} and \eqref{eq:Rbudget_Gama}.  We first note that  \eqref{eq:Acc-mcnn-Gama} is monotonically increasing in function of $R_\text{L}$ and $R_\text{H}$, since $A_{3D}>A_{2D}$ and $A_{2D}>A_{SP}$. This allows for the use numerical methods that gradually explore the parameter space of  $\{R_{\text{L}}, R_{\text{H}}\}$ by setting $R_{\text{sent}}$ in  \eqref{eq:Rbudget_Gama} as close as possible to $R_{\text{available}}$ and then finding the maximum values for   $\{R_{\text{L}}, R_{\text{H}}\}$ that satisfy  \eqref{eq:Rbudget_Gama}, since such values will automatically maximize   \eqref{eq:Acc-mcnn-Gama}. 

In our experiments, amongst several alternatives, we opted for the method of Toint \textit{et al.}
\cite{conn1997globally}, which finds the  solution $\{R^*_{\text{L}}, R^*_{\text{H}}\}$  that maximizes   \eqref{eq:Acc-mcnn-Gama}  under the constraint  $R_{\text{sent}}\leq R_{\text{available}}$   with the provision of sufficient exploration time. Given that this optimization process is done offline based on training data $\mathcal{V}$, this does not impose any overhead at runtime. Finally, we remark that, in case
$R_\text{motion}$ is not measurable at training or test time, the optimization method proposed in this section can be generalized to other features that correlate with $R_{\text{motion}}$   (e.g.  number of MVs per frame).

\section{Experimental Results}\label{sec:Evaluation}



\subsection{Used Datasets and Rate Saving from Cropped Bitstreams}

We train and test our  2D and 3D CNN architectures
on eight distinct motion vector datasets generated by varying the QP setting of  
AVC/H.264  and HEVC to encode UCF-101\cite{soomro2012ucf101}, while skipping texture information
as described in Section \ref{sec:highly_compressed_bitstreams}. For all videos:
the first frame is encoded
as an IDR (with remaining frames inter-predicted  as P-frames), the frame rate is set to
$25$, and we set the motion vector search
range to $16$ pixels. Since specifying a particular  quantization parameter
has a direct effect on  the MVs produced by  AVC/H.264 and HEVC, this  gives several
distinct source distributions for the classifier to be trained and tested
on.

\begin{table}[t] \caption{Average AVC/H.264 bitrate (kbps) of  UCF-101; $R_{\text{orig}}$
is the bitrate of the original bitstream, $R_{\text{cropped}}$ is the bitrate
 after cropping and retaining texture and motion information, and $R_\text{motion}$
is the MV bitrate.} \centering \begin{footnotesize}
\begin{tabular}{cccccc} 
\toprule
& \multicolumn{3}{c}{}&    \multicolumn{2}{c}{\% of $R_{\text{motion}}$ to}
\\
QP & $R_\text{orig}$   & $R_\text{cropped}$ & $R_\text{motion}$ & $R_\text{orig}$
& $R_\text{cropped}$\\  \midrule
0 & 4273.0 & 321.3 & 155.4 & 3.6 & 48.3\\
30 & 274.9 & 112.3 & 46.9 & 17.0 & 41.7 \\
40 & 80.0 & 49.9 & 18.5 & 23.2 & 37.1\\
51 & 27.7 & 20.0 & 4.6 & 16.7 & 23.1\\ \bottomrule 
 \end{tabular} \end{footnotesize}   \label{tab:jm_rate_A} \end{table}

\vspace*{-1mm}

\begin{table}[t] \caption{Average HEVC   bitrate (kbps) of  UCF-101; $R_{\text{orig}}$
is the bitrate of the original bitstream, $R_{\text{cropped}}$ is the bitrate
 after cropping and retaining texture and motion information, and $R_\text{motion}$
is the MV bitrate.} \centering \begin{footnotesize}
\begin{tabular}{cccccc} \toprule
& \multicolumn{3}{c}{}&    \multicolumn{2}{c}{\% of $R_{\text{motion}}$ to}
\\
QP & $R_\text{orig}$   & $R_\text{cropped}$ & $R_\text{motion}$ & $R_\text{orig}$
& $R_\text{cropped}$\\  \midrule
0 & 3065.2 & 204.9 & 39.9 & 1.3 & 19.1\\
30 & 157.7 & 58.8 & 12.0 & 7.6 & 20.6 \\
40 & 40.2 & 26.7 & 4.9 & 2.5  & 12.25\\
51 & 10.9 & 9.8 & 0.8 & 7.3 & 8.1\\ \bottomrule 
 \end{tabular} \end{footnotesize}   \label{tab:hm_rate_A} \end{table}


%
%
%
%
%
%
%
%
%
%
%
%
%
%
%
%


\begin{figure}[ht]\hspace*{-5mm}   \centering \includegraphics[scale=0.57]{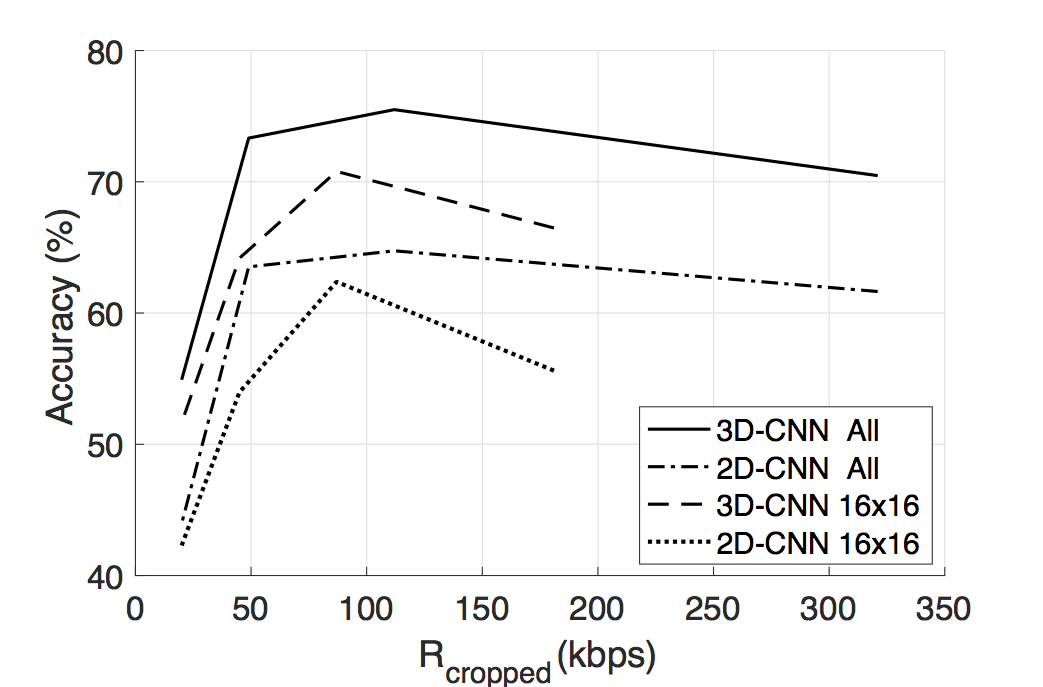}
        \caption{\label{fig:rate_accuracy_JM} Rate-accuracy after cropped
AVC/H.264  bitstreams are passed  to the 2D and 3D classifiers. Each point
for every curve corresponds to a different QP setting during encoding, with
``$16\times16$'' indicating  restriction to  $16\times16 $ blocks (no
MB\ subblocks) and ``All'' indicating the use of all MB partitions. } \end{figure}

\begin{figure}[ht] \hspace*{-4mm} \centering \includegraphics[scale=0.58]{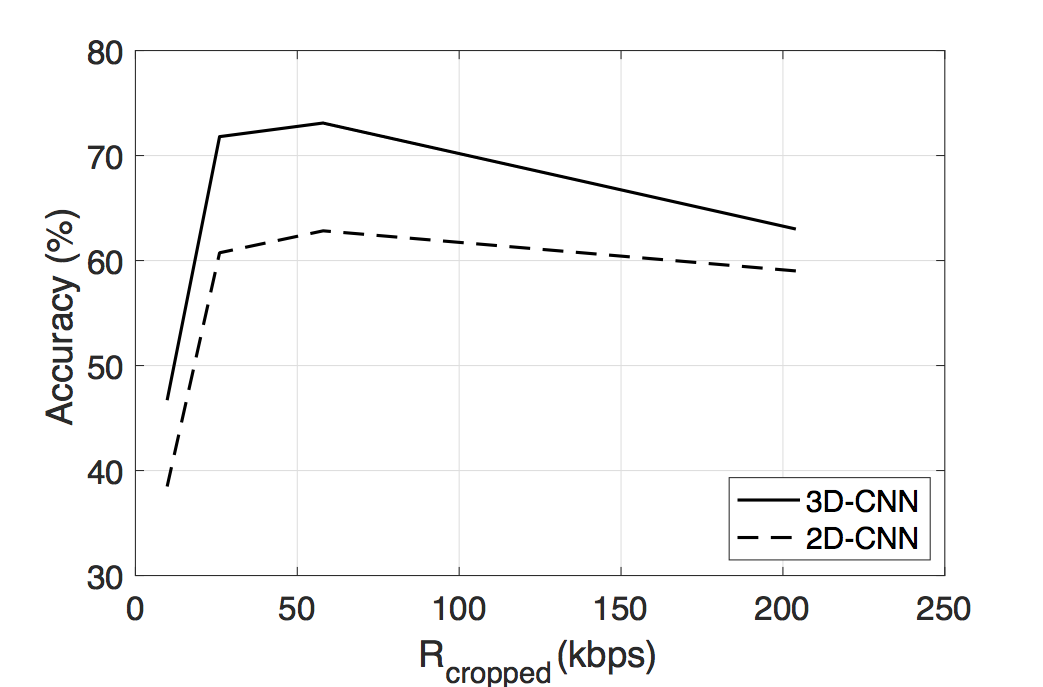}
        \caption{\label{fig:rate_accuracy_HM} Rate-accuracy after cropped
HEVC  bitstreams are passed  to the 2D and 3D temporal CNNs. Each point
for every curve corresponds to a different QP setting during encoding, with $\text{encoder
parameter CBT Depth} = 2$. } \end{figure}

\begin{figure}[ht!] \hspace*{-4mm} \centering \includegraphics[scale=0.56]{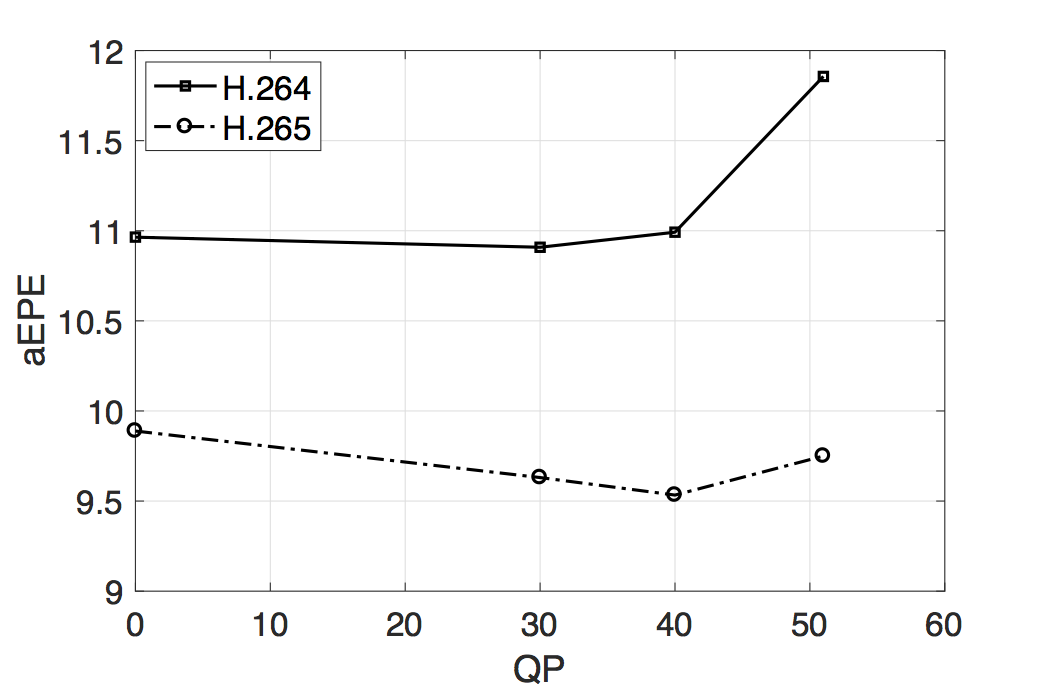}
        \caption{\label{fig:aEPE} Average EPE between our approximated optical flow with different QP settings and an estimated dense optical flow  ground truth using   the method of Brox \textit{et al.} \protect\cite{brox2011large}. } \end{figure}

\begin{table}[ht] 

\centering

\begin{tabular}{ccccccc}
 
\toprule 

 Framework & $R_{\text{cropped}}$ &
\multicolumn{2}{c}{Accuracy (\%)} \\

& (kbps) & \begin{scriptsize}UCF\end{scriptsize} & \begin{scriptsize}HMDB\end{scriptsize}
  \\ \midrule

 3D-CNN-F (H.264, $\text{QP}=30$) &  112.3 & 88.1 & 53.0 \    \\

 {3D-CNN-F (H.264, $\text{QP}=40$)} &   49.9 & 88.1 &  52.9\    \\
 
 {3D-CNN-F (H.264, $\text{QP}=51$)}  &  20.0 &  84.0 & 47.7  \\
 
 \midrule

 3D-CNN-F (H.265, $\text{QP}=30$) &  58.8  & 86.7 & 50.9     \\
 
{3D-CNN-F (H.265, $\text{QP}=40$)}  &  26.7 & 86.6 &  50.7       \\
 
 {3D-CNN-F (H.265, $\text{QP}=51$)} &  9.8  & 81.4 & 47.1 \\
 
\midrule

EMV + RGB-CNN \cite{zhang2016real} & --- & 86.4 & ---\\
MVCNN \cite{chadha2017video} & --- & 89.8 & 56.0 \\

CoViAR \cite{wu2017compressed} & --- & 90.4 & 59.1 \\
 ST-ResNet + iDT \cite{feichtenhofer2016convolutional}  & --- & 94.6 & 70.3 \\

 ActionVLAD + iDT \cite{feichtenhofer2016convolutional} 
& --- & 93.6 & 69.8 \\

TSN (3 modalities)\cite{wang2016temporal} & --- & 94.2 & 69.4 \\
I3D\cite{carreira2017quo} & ---  &  93.4 & 66.4 \\
TSCNN (SVM fusion) \cite{simonyan2014two} & --- &  88.0 & 59.4 \\
LTC\cite{varol2016long} & ---  &  91.7 & 64.8\ \\
C3D (3 nets)+IDT\cite{tran2015learning}& ---  &  90.4 & --- \\
\bottomrule 
\end{tabular}
	\caption{\label{tab:two_stream} Comparison of   our 3D-CNN-F classifier (fusion of VGG-16 spatial CNN and 3D-CNN as shown in Fig. \ref{fig:mcnn}) against state-of-the-art  CNNs.}

\end{table}

\subsection{Rate-Accuracy Results}             
            
As the quality of predictions made by  CNN models is strongly
tied to the properties of the source distribution (e.g. cross-class variance,
noise), we expect that varying the rate should affect  the
 accuracy of our classifier accordingly.
Since the QP\ values control the video rate, we first show visual examples of the effect of QP on the quality of  approximated sparse optical flow in Fig. \ref{fig:mix_sparsity}. The best approximations appear to be for QP\ values in the region of $30$ to $40$. To assess the rate savings and classification accuracy of our proposal when varying QP values, in Table \ref{tab:jm_rate_A} and Table \ref{tab:hm_rate_A} we compare the original bitrate, $R_{\text{orig}}$, with the bitrate of the cropped bitstreams, $R_{\text{cropped}}$, and the rate of  retained motion vectors,    $R_{\text{motion}}$. The results show  that streaming cropped bitstreams allows
for 28\% to 92\% reduction in bitrate for AVC/H.264,  and 11\% to 94\% for HEVC. The related classification accuracy results are presented in Fig. \ref{fig:rate_accuracy_JM} and Fig. \ref{fig:rate_accuracy_HM}. As indicated by the visual examples of Fig. \ref{fig:mix_sparsity},  the utilized CNNs indeed achieve their best accuracies
at QP\ values of $30$ to $40$. 

Importantly, we  observe that rate-accuracy curves are not monotonic
(i.e., accuracy decreases for very low or very high QP\ values). 
We  expect sparser motion vectors (e.g., MVs produced by setting $\text{QP}=51$ where the rate allocated to motion vectors is the lowest) to make certain 
classes with high motion similarity particulary harder to classify and easier
to confuse with each other.
On the other hand, as shown by Fig. \ref{fig:mix_sparsity}, setting  $\text{QP}<30$   also has
a detrimental effect on accuracy, since the derived MVs become significantly
more noisy due to the inadequacy of the simple translational block model
of AVC/H.264 and HEVC to smoothly approximate the optical flow field since such block models are optimized for rate control and not optical flow estimation
\cite{chadha2017video,brox2011large}. 

To cross validate  with an external benchmark, Fig. \ref{fig:aEPE} shows the  average End Point Error (aEPE) between MV frames and a dense optical flow ground truth approximated using the method proposed  by Brox \textit{et al.}  \cite{brox}. The resulting  curves show that, for both video coders, the minimum aEPE value against dense optical flow is in the QP range of $30$ to $40$.  We also note that the best performance occurs at a  lower rate for HEVC compared to AVC/H.264, which is due to the  enhanced coding efficiency and improved inter-frame macroblock search of the  HEVC\ standard. This is also reflected in  Fig. \ref{fig:aEPE}, where the aEPE of  HEVC\ is lower than that of AVC/H.264 over all QP settings.     
\subsection{Comparison Against External Benchmarks}

In Table \ref{tab:two_stream}, we report the accuracy of our fused spatio-temporal classifier of Fig. \ref{fig:mcnn}, wherein the predictions of the spatial and temporal classifiers are averaged, and compare against state-of-the-art methods from the literature. Our results
show that our approach remains competitive to the state-of-the-art on UCF-101,
 while  retaining  the significant bitrate gains reported in Table \ref{tab:jm_rate_A} and Table \ref{tab:hm_rate_A}.
In addition, while our approach is outperformed   by methods like ST-ResNet and TSN,
it is important to emphasize that these methods are orders-of-magnitude more complex than  operating with sparse
compressed-domain
information  \cite{girdhar2017actionvlad,chadha2017video,zhang2016real}, since they  require the
use of dense optical flow and need to receive
and decode 
entire video bitstreams.  Moreover, ST-ResNet and TSN use significantly deeper neural network architectures in comparison to our approach, which makes their inference  significantly more compute intensive than the CNN architectures of Fig. \ref{fig:mcnn}. Finally, in order to improve our results for the HMDB dataset, our rate-optimization method can be applied in conjunction with the recent motion vector accumulation method proposed in CoViAR \cite{wu2017compressed}, which uses compressed-domain information to infer a sparse optical flow representation. While their  optical flow approximation method is  more complex in comparison to ours, by applying our  classifier selection framework to such representations  it is possible to gain  even more savings in bitrate.

\subsection{MCNN  Performance }\label{sec:mcnn_performance}

\begin{figure}[ht!]\hspace*{-4.0mm} \centering \includegraphics[scale=0.66]{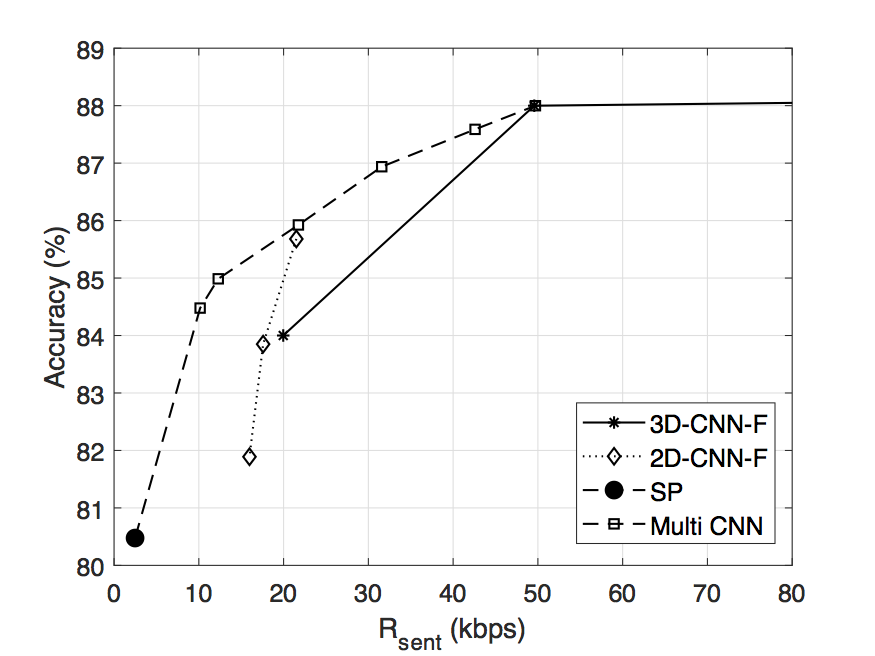}
        \caption{\label{fig:multicnn_rate_accuracy} Rate-accuracy  results on the UCF-101 dataset.  For the 3D-CNN-F   and  2D-CNN-F  classifiers (fusion of spatial CNN with 3D/2D motion CNNs as shown in Fig. \ref{fig:mcnn}), different rates are obtained by using different QP settings.
When using Multi-CNN,\ rate is controlled by setting $\text{QP}=40$ and varying $R_{\text{available}}$
to solve for $R^*_\text{L}$ and $R^*_\text{H}$. Note that the leftmost point shows the performance when the temporal stream is not used and the  MCNN selector only considers the outputs of the spatial stream model. 
} \end{figure}

To study the  performance of our proposed  MCNN under varying rate constraints, we solve \eqref{eq:Maxmin}
 for multiple values of $R_{\text{available}}$ within the interval $[0,50]$ kbps as described in Section \ref{sec:OptMultiCNN}. We then assess the MCNN accuracy on the UCF-101 test set for each set of parameters $\{R^*_{\text{L}},
R^*_{\text{H}}\}$
and show the results in Fig. \ref{fig:multicnn_rate_accuracy}. When using the optimization framework of  Section \ref{sec:OptMultiCNN}, 
 approximately 25 kbps (50\%) reduction in bitrate can be obtained against the 3D-CNN-F  classifier (25 kbps vs. 50 kbps)  at less than 2\% reduction in classification accuracy.  Importantly,
   further bitrate reductions  are made possible with graceful (and monotonic) degradation
in classification accuracy, to the point of making it viable to get an accuracy within 7\% from the top performance at an average bitrate as low as $R_\text{sent}=3$ kbps. This shows the potential
for further exploration of rate-accuracy optimization in CNN-based video
classification and the utility of features such as $R_{\text{motion}}$ in inferring the temporal information needed for classification. \\


%

%
%

\section{Conclusion}\label{sec:Conclusion}
We present the first exploration of rate-accuracy trade-offs in advanced
video classification with CNNs. Given that our proposed method can be applied based on standardized codecs with minimal bitstream modifications, it is well suited for   visual IoT\ or semantic video crawling applications. The obtained results show that, when reducing bitstreams to the necessary elements for 2D\ or 3D\ CNN classification,    28\%-92\% and 11\%-94\% reduction in bitrate can be achieved for
AVC/H.264 and HEVC respectively. We have observed that  non-monotonic rate-accuracy curves are obtained by state-of-the-art CNNs classifying approximated flow from compressed bitstreams (following the AVC/H.264 and HEVC standards). On the other hand, a rate-based selection
method between  multiple CNN\ classifiers with  varied
 input  requirements is shown to achieve monotonic rate-accuracy characteristics and allow for even further rate gains, with
minimal impact on classification accuracy.  Our implementation and all tools needed to reproduce our results are available online at: \href{https://github.com/rate-accuracy-mvcnn/main}{https://github.com/rate-accuracy-mvcnn/main}.

\appendix

We validate our modelling choices described in Section \ref{sec:OptMultiCNN}.
For brevity of exposition, all figures and results here are reported for the indicative case of AVC/H.264 with  $\text{QP}=40$.

\subsection{Distribution of $R_\text{motion}$ and Performance Overlap \label{sec:app_fitted_model_perf_overlap}}

\begin{figure}[ht] \centering \includegraphics[scale=0.5]{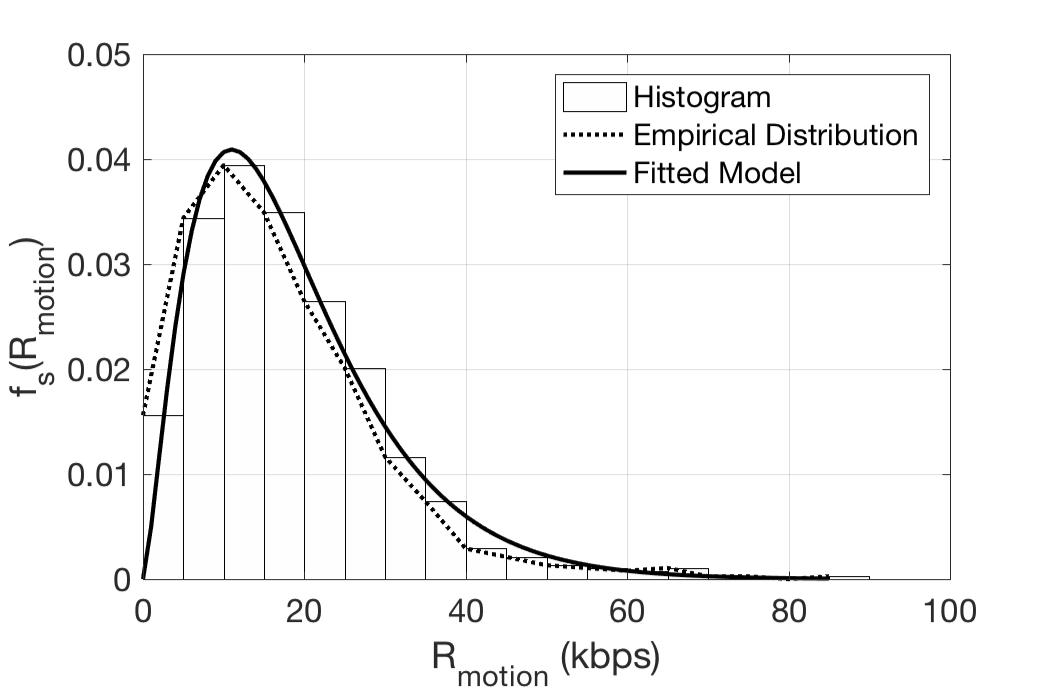}
        \caption{\label{fig:rmotion_gamma} Empirically measured distribution
of $R_{\text{motion}}$ and  fitted Gamma distribution with shape and scale parameters: $\alpha=2.43$, $\beta=0.13$.}
\end{figure}

\begin{figure}[ht] \centering \includegraphics[scale=0.49]{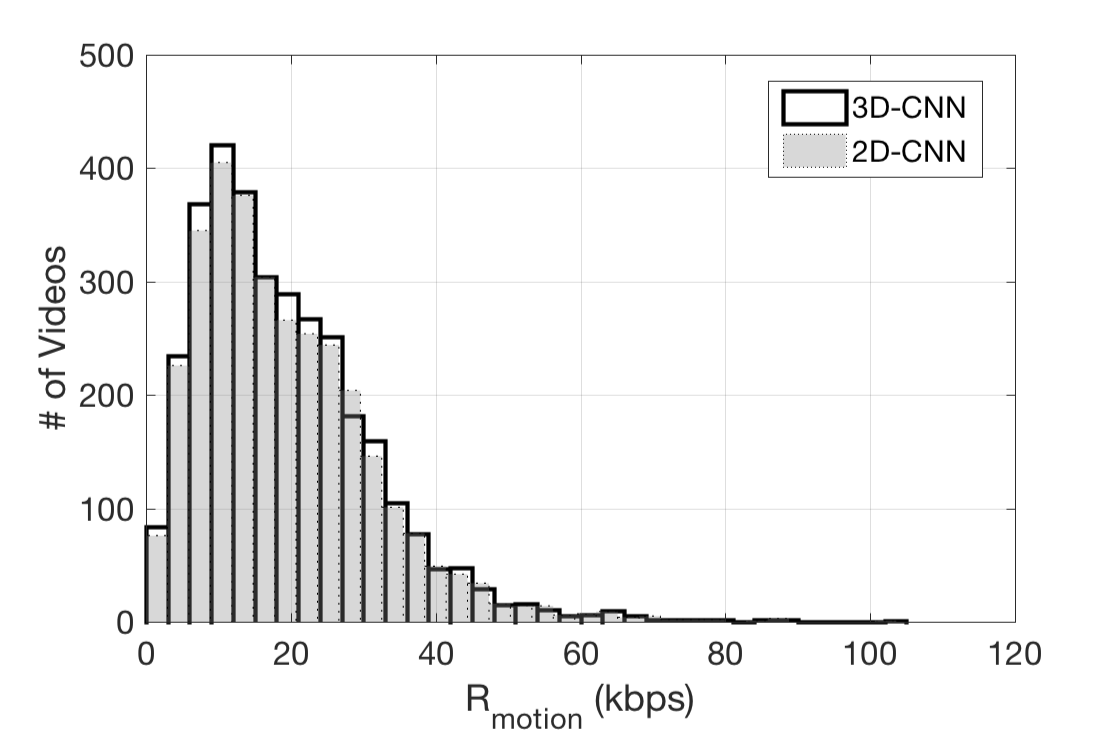}
        \caption{\label{fig:performance_overlap} Number of videos classified correctly by each temporal CNN classifier for different values of  $R_{\text{motion}}$. } \end{figure}

 In this section we  compare the distribution
of $R_{\text{motion}}$ against the fitted model and verify  the overlap of performance between the proposed architectures in Section \ref{sec:network_design}.  All of the UCF-101 dataset is used to produce the results shown in Fig. \ref{fig:rmotion_gamma} and Fig. \ref{fig:performance_overlap}.
For  Fig. \ref{fig:rmotion_gamma}, the Kullback-Leibler divergence (describing the distance between the empirical and fitted Gamma distribution) was found to be $0.034$. This proximity justifies our use of this distribution for characterizing the probability of occurrence of  different values of $R_{\text{motion}}$.
Concerning  Fig. \ref{fig:performance_overlap}, the experiments show that the 3D and 2D CNN architectures perform similarly for middle-range values of $R_{\text{motion}}$, with the 3D-CNN outperforming the 2D-CNN for most of the lower MV bitrates. The performance of both CNNs decays for high values of $R_{\text{motion}}$. Hence, for the high-end range of $R_{\text{motion}}$,  only the spatial CNN should be used (VGG-16 of  Fig. \ref{fig:mcnn}).   

\subsection{Linear Model Verification for \eqref{eq:R_3D_2D_SP} \label{sec:app_3d_2d}}

\begin{figure}[h!] \centering \includegraphics[scale=0.52]{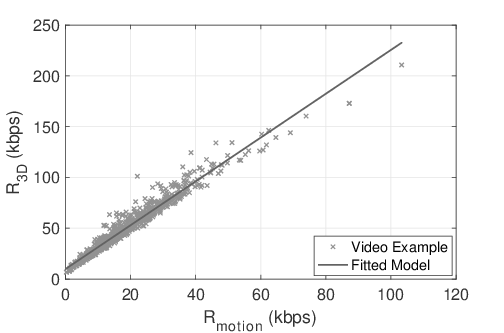}
        \caption{\label{fig:r3d_rmotion} Bitrate of inputs sent to 3D architecture
$R_{\text{3D}}$ plotted against
$R_{\text{motion}}$ and fitted model of $R_{\text{3D}}$
with linear coefficients $a_{\text{3D}}=2.21$ and $b_{\text{3D}}=9.04$.} \end{figure}

\begin{figure}[h!] \centering \includegraphics[scale=0.52]{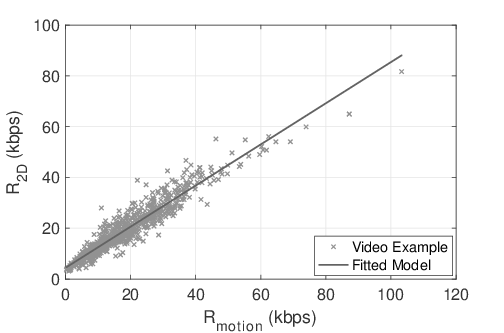}
        \caption{\label{fig:r2d_rmotion} Bitrate of inputs sent  to 2D architecture
$R_{2D}$ plotted against 
$R_{\text{motion}}$ and fitted model of $R_{\text{2D}}$
with linear coefficients $a_{\text{2D}}=0.83$ and $b_{\text{2D}}=4.27$.} \end{figure}

We selected  5\% of the\  UCF-101 videos randomly and present the plots of  $R_{\text{motion}}$ vs. $R_{\text{3D}}$ and  $R_{2D}$ in Fig \ref{fig:r3d_rmotion} and Fig.  \ref{fig:r2d_rmotion} . Using the same set, we  calculated the  coefficient of determination $R^2$
 to relate the experimental variance  to the residual variance of  the linear model
and found it to be 93\% for $R_{\text{3D}}$ and 88\% for $R_{2D}$. Similar results have been obtained for the HMDB\ dataset. These results validate that the linear assumption of  \eqref{eq:R_3D_2D_SP} is a good approximation.

{\small
\bibliographystyle{IEEEtran}
\bibliography{literature}
}

\end{document}